\documentclass[runningheads]{llncs}
\usepackage{graphicx}
\usepackage{amsmath}
\usepackage{amsfonts}
\usepackage{multirow}
\usepackage{times}
\usepackage{float}
\usepackage{booktabs}
\usepackage{algorithmic}
\usepackage{color}
\usepackage[colorlinks]{hyperref}

\newcommand{\bheading}[1]{{\noindent{\textbf{#1}}}}

\begin{document}

\title{A Hierarchical Feature Constraint to Camouflage Medical Adversarial Attacks}

\author{Qingsong Yao\inst{1} \and
Zecheng He\inst{2} \and
Yi Lin\inst{3} \and
Kai Ma\inst{3} \and
Yefeng Zheng\inst{3} \and
S. Kevin Zhou\inst{1}}



\institute{\
Medical Imaging, Robotics, Analytic Computing Laboratory/Engineering (MIRACLE), Key Lab of Intelligent Information Processing of Chinese Academy of Sciences (CAS), Institute of Computing Technology, CAS, Beijing 100190, China\\
 \email{\{yaoqingsong19\}@mails.ucas.edu.cn} \email{\{s.kevin.zhou\}@gmail.com}  \and
Princeton University \\ \email{zechengh@princeton.edu} \and
Tencent Jarvis Lab}

\authorrunning{Q. Yao, et al.}
\maketitle

\begin{abstract}
Deep neural networks for medical images are extremely vulnerable to adversarial examples (AEs), which poses security concerns on clinical decision-making. Recent findings have shown that existing medical AEs are easy to detect in feature space. To better understand this phenomenon, we thoroughly investigate the characteristic of traditional medical AEs in feature space. Specifically, we first perform a stress test to reveal the vulnerability of medical images and compare them to natural images. Then, we theoretically prove that the existing adversarial attacks manipulate the prediction by continuously optimizing the vulnerable representations in a fixed direction, leading to outlier representations in feature space. Interestingly, we find this vulnerability is a double-edged sword that can be exploited to help hide AEs in the feature space. We propose a novel hierarchical feature constraint (HFC) as an add-on to existing white-box attacks, which encourages hiding the adversarial representation in the normal feature distribution. We evaluate the proposed method on two public medical image datasets, namely {Fundoscopy} and {Chest X-Ray}. Experimental results demonstrate the superiority of our HFC as it bypasses an array of state-of-the-art adversarial medical AEs detector more efficiently than competing adaptive attacks.
\end{abstract}

\section{Introduction}

Deep neural networks (DNNs) are vulnerable to adversarial examples (AEs)~\cite{szegedy2013intriguing}. AEs are maliciously generated by adding human-imperceptible perturbations to clean examples, compromising a network to produce the attacker-desired incorrect predictions~\cite{dong2018boosting}. The adversarial attack in medical image analysis is disastrous as it can manipulate patients' disease diagnosis and cause serious subsequent problems. More disturbingly, recent studies have shown that DNNs for medical image analysis, including disease diagnosis~\cite{paschali2018generalizability,finlayson2018adversarial}, organ segmentation~\cite{ozbulak2019impact}, and landmark detection~\cite{yao2020miss}, are more vulnerable to AEs than natural images. 

\begin{figure}[t]
\begin{center}
   \includegraphics[width=0.6\linewidth]{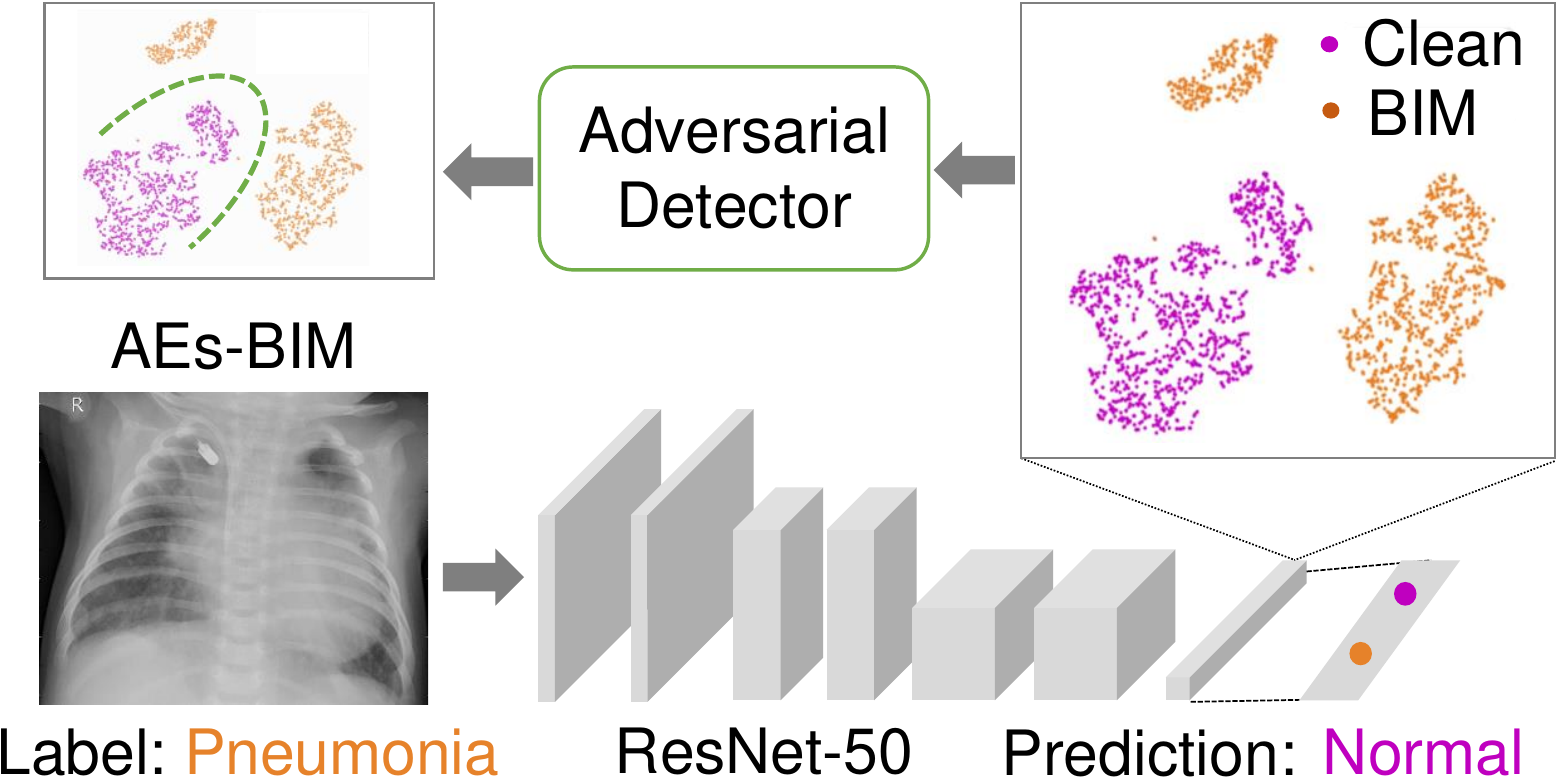}
\end{center}
   \caption{We craft AEs by the iterative basic method (BIM)~\cite{bim} to manipulate the prediction and visualize the penultimate layer's features of the adversarial and clean examples by 2D t-SNE.}
\label{Fig:Main}
\end{figure}

On the other hand, recent defenses~\cite{ma2020understanding,li2020robust} have shown that medical AEs can be easily detected in feature space. We plot the 2D t-SNE~\cite{t-SNE} in Fig.~\ref{Fig:Main} to illustrate the differences between clean and adversarial features from the penultimate layer of a well-trained pneumonia classifier, revealing that adversarial attacks move the deep representations from the original distribution to extreme outlier positions. As a result, a defender can easily take advantage of this characteristic to distinguish AEs. 

Given this phenomenon, two key questions are investigated in this paper. The first one is:
\textit{What causes medical AEs easier to be detected, compared to AEs of natural images?} To better understand this, we conduct both empirical and theoretical analyses. Firstly, we discover that medical features are more vulnerable than natural ones in a stress test, which aims to change the features by the adversarial attack. Then, we theoretically prove that to generate medical AEs, the adversarial features are pushed in a consistent direction towards outlier regions in the feature space where the clean features rarely reside. In consequence, medical adversarial features become outliers. 

The second question is: \textit{Can we hide a medical AE from being spotted in the feature space?} A representative adaptive attack selects a guide example and forces the AE feature to be close to the guide feature~\cite{feature_iclr}. However, this does not directly apply to medical AES. Different medical images have different backgrounds and lesions, making it difficult to manipulate the AE features to be the same as the guide one in all layers within limited perturbation. To hide the adversarial representation at a low price, we propose a novel hierarchical feature constraint (HFC) as an add-on term, which can be plugged into \textit{existing white-box attacks}. HFC first models the normal feature distributions for each layer with a Gaussian mixture model, then encourages the AEs to move to the high-density area by maximizing the log-likelihood of the AE features. 

We perform extensive experiments on two public medical diagnosis datasets to validate the effectiveness of HFC. HFC helps medical AE bypass state-of-the-art adversarial detectors, while keeping the perturbations small and unrecognizable to humans. Furthermore, we demonstrate that HFC significantly outperforms other adaptive attacks~\cite{feature_iclr,CW_bpda,CW_ten} on manipulating adversarial features. Our experiments support that \textbf{the greater vulnerability of medical representations allows an attacker more room for malicious manipulation}. 

Overall, we highlight the following contributions:
\begin{itemize}
    \item We investigate the feature space of medical images and shed light on why medical AEs can be more easily detected, compared with those of natural images.
    \item We propose a hierarchical feature constraint (HFC), a novel plug-in that can be applied to existing white-box attacks to avoid being detected.
    \item Extensive experiments validate the effectiveness of our HFC to help existing adversarial attacks bypass state-of-the-art adversarial detectors simultaneously with small perturbations.
\end{itemize}

\section{Related Work}

\bheading{Preliminaries.} Given a clean image $x$ with its label $y \in [1,2,\ldots,Y]$ and a DNN classifier $h$ 
, the classifier predicts the class of the input example $y'$ via:
\begin{align}
    y' &= \mathop{\arg\max}\limits_k~p(k|x) \equiv \frac{\mathop{\exp}(l_k(x))} {\sum_{j=1}^K\mathop{\exp}(l_{j}(x))}.
\label{Eq:inference}
\end{align}
where the logits  $l_k(x)$ (with respect to class $k$) is given as $l_k(x)= \sum_{n=1}^N{w_{nk}*z_n(x)} + b_k$, in which $z_n(x)$ is the $n^{th}$ activation of the penultimate layer that has $N$ dimensions; $w_{nk}$ and $b_k$ are the weights and the bias from the final dense layer, respectively.

\bheading{Adversarial attack.} A common way is to manipulate the classifier's prediction by minimizing the classification error between the prediction and target class $c$, while keeping the AE $x_{adv}$ within a small $\epsilon$-ball of the $L_p$-norm~\cite{PGD} centered at the original sample $x$, i.e., $\|x_{adv} - x\|_p \leq \epsilon$.
In this paper, we focus on typical $L_\infty$ adversarial attacks, which are most commonly used due to its consistency with respect to human perception~\cite{PGD}. The existing $L_\infty$ approaches can be categorized into two parts. The first one is gradient-based approaches, such as the fast gradient sign method (FGSM)~\cite{goodfellow2014explaining}, basic iterative method (BIM)~\cite{bim}, the momentum iterative method (MIM)~\cite{dong2018boosting} and projected gradient descent (PGD)~\cite{PGD}. BIM computes AE as follows: 
\begin{align}
 x_0^*=x; \qquad x_{t+1}^* = \Pi_\epsilon( x_t^* - \alpha \cdot \mathrm{sign}(\nabla_x J(h(x_t^*), c))).
\label{Eq:BIM}
\end{align}
where $J$ is often chosen as the cross-entropy loss; $\epsilon$ is the $L_\infty$ norm bound; $\alpha$ is the step size; and $\Pi(\cdot)$ is the project function. Differently from BIM, PGD use a random start $x_0 = x + U^d(-\epsilon, \epsilon)$, where $U^d(-\epsilon,\epsilon)$ is the uniform noise between $-\epsilon$ and $\epsilon$.

The second category is optimization-based methods, among which one of the representative approach is the Carlini and Wanger (CW) attack~\cite{cwattack}. According to~\cite{PGD}, the $L_\infty$ version of CW attack can be solved by using the following objective function:
\begin{align}
   \hat{J} = \max(l_{c}(x_t^*) - l_{{y_{max} \neq c}}(x_t^*), -\kappa).
\label{Eq:CW}
\end{align}
where $l_{c}$ is the logits with respect to the target class; $l_{y_{max} \neq c}$ is the maximum logits of the remaining classes; and $\kappa$ is a parameter managing the confidence.



\bheading{Adversarial defenses.} Various proactive defenses~\cite{taghanaki2019kernelized,he2019non} have been proposed to defend adversarial attacks, such as feature squeezing~\cite{xu2017feature}, distillation network~\cite{papernot2016distillation}, JPEG compression~\cite{jpeg}, gradient masking~\cite{masking}. Per~\cite{dong2019benchmarking}, the adversarial training~\cite{goodfellow2014explaining,tramer2017ensemble,PGD} is the most robust defense, which augments the training set with AEs but consumes too much training time. However, these defenses can be bypassed either completely or partially by adaptive attacks~\cite{CW_bpda,CW_ten,tramer2020adaptive}. 
Different from the challenging proactive defense, reactive defenses have been developed to detect AEs from clean examples with high accuracy~\cite{zheng2018robust}. For example, learning-based methods (e.g., RBF-SVM~\cite{SaftyNet}, DNN~\cite{metzen2017detecting}) train a decision boundary between clean and adversarial features,
while anomaly-detection based methods directly reject the outlier feature of the AEs. Specifically, kernel density estimation (KDE)~\cite{kde} and multivariate Gaussian model (MGM)~\cite{li2020robust} modeled the normal feature distribution. Local intrinsic dimensionality (LID)~\cite{ma2018characterizing} characterized the dimensional properties of the adversarial subspaces. The degree of the outlier was measured by  Mahalanobis distance (MAHA)~\cite{MAHA}. Especially, medical AEs were proven to be easy to detect with nearly \textit{100\%} accuracy by these approaches~\cite{ma2018characterizing,li2020robust}.

\section{Why are Medical AEs Easy to Detect?} \label{sec:why}



\subsubsection{Vulnerability of representations}\label{sec:why:vulnerability}
A stress test is performed to evaluate the vulnerability of the features of medical and natural images. Specifically, we aim to manipulate the features as much as possible by adversarial attacks. In the test, we decrease ($\downarrow$) and increase ($\uparrow$) the features by replacing the loss function $J$ in BIM (Eq.~\ref{Eq:BIM}) by $J_\downarrow^* = \mathrm{mean}(f^l(x))$ and $J_\uparrow^* = - \mathrm{mean}(f^l(x)),$ respectively, where $f^l(x)$ is the feature from the $l^{th}$ activation layer. We execute the stress test on the medical image dataset (Fundoscopy~\cite{aptos}) and natural image dataset (CIFAR-10~\cite{cifar}). The results in Table~\ref{Table:mean} demonstrate that the features of the medical images can be altered more drastically, i.e., \textbf{the medical image representations are much more vulnerable than natural images.}

\begin{table}[htp]
\caption{Comparison of the robustness between medical images and natural images. We calculate the mean values of the activation layers from ResNet-50 before and after the adversarial attacks. The stress test uses BIM with perturbations under the constraint $L_\infty=1/256$. }
\centering
\begin{tabular}{l|rrr|rrr}
\bottomrule \hline
Dataset            & \multicolumn{3}{c|}{Fundoscopy} & \multicolumn{3}{c}{CIFAR-10} \\
\hline
Layer index & 36        & 45       & 48      & 36       & 45      & 48      \\
\hline
Normal                     & .0475    & .1910   & .3750   & .0366   & .1660   & .1900    \\
Adversarial ($\downarrow$) & .0322    & .0839   & .0980   & .0312   & .1360   & .1480   \\
Adversarial ($\uparrow$)   & .0842    & .7306   & 2.0480  & .0432   & .2030   & .2640 \\
Difference ($\downarrow$)  & .0153    & .1071   & .2770   & .0054   & .0300   & .0420 \\
Difference ($\uparrow$)    & .0367    & .5396   & 1.6730  & .0066   & .0370   & .0740  \\
\hline \toprule
\end{tabular}
\label{Table:mean}
\end{table}

\subsubsection{Consistency of gradient direction}

To investigate how the feature vulnerability can be exploited by adversarial attacks, we then focus on the loss function and the corresponding gradient on the final logits output $l_k(x)$. In each iteration of the $L_\infty$ attack introduced above, $J$ and $\hat{J}$ increase the logits of the target class and decrease the others simultaneously. As a result, gradients point to a similar direction across iteration, which will be back-propagated according to the chain rule.

\bheading{Theorem 1.} \textit{Consider a binary disease diagnosis network and its representations from the penultimate layer, the directions of the corresponding gradients are fixed during each iteration under adversarial attack.}\footnote{We provide the proof in the supplementary material.}


\begin{figure}[h]
\begin{center}
  \includegraphics[width=0.27\linewidth]{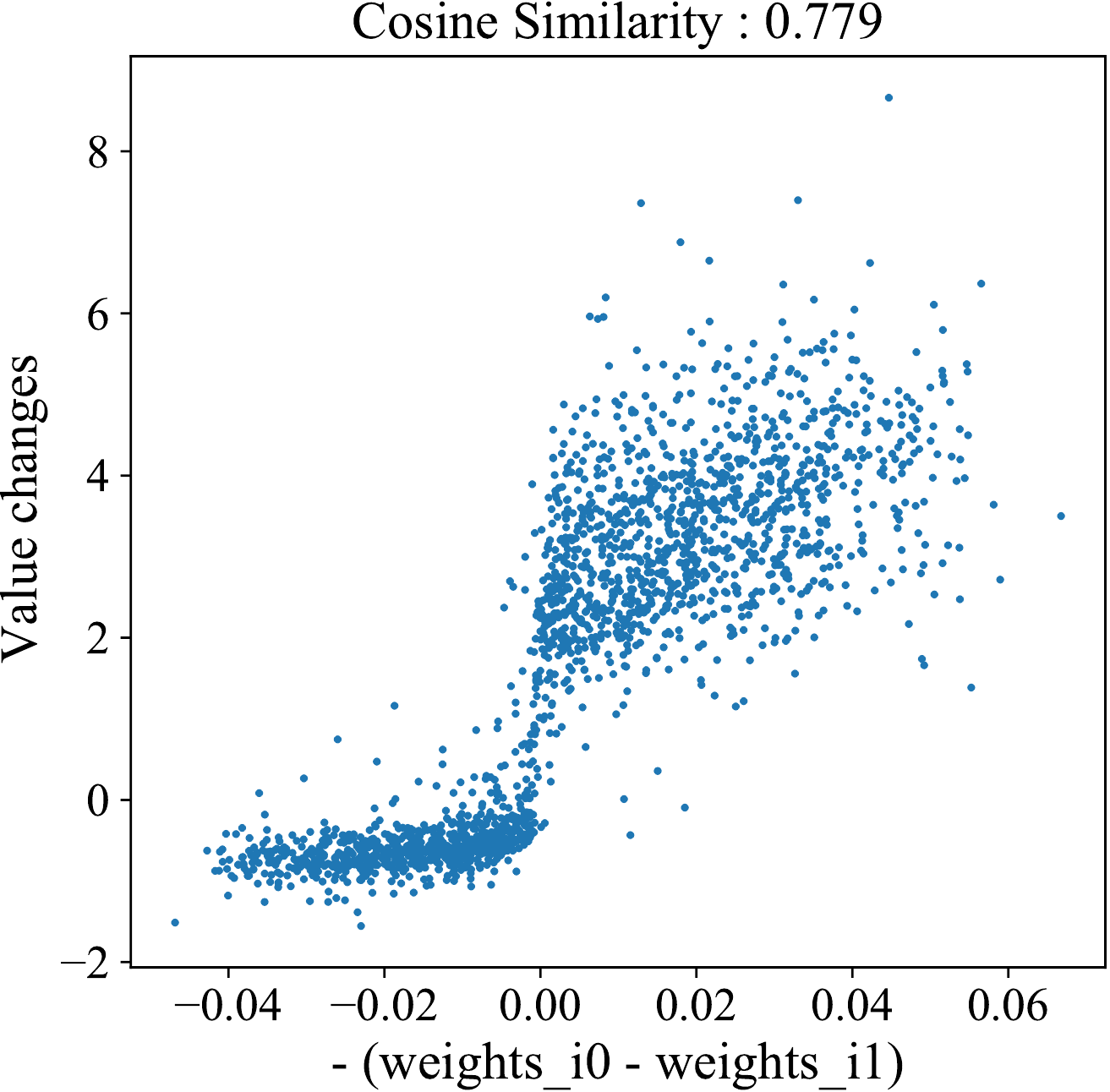}(a)
     \includegraphics[width=0.6\linewidth]{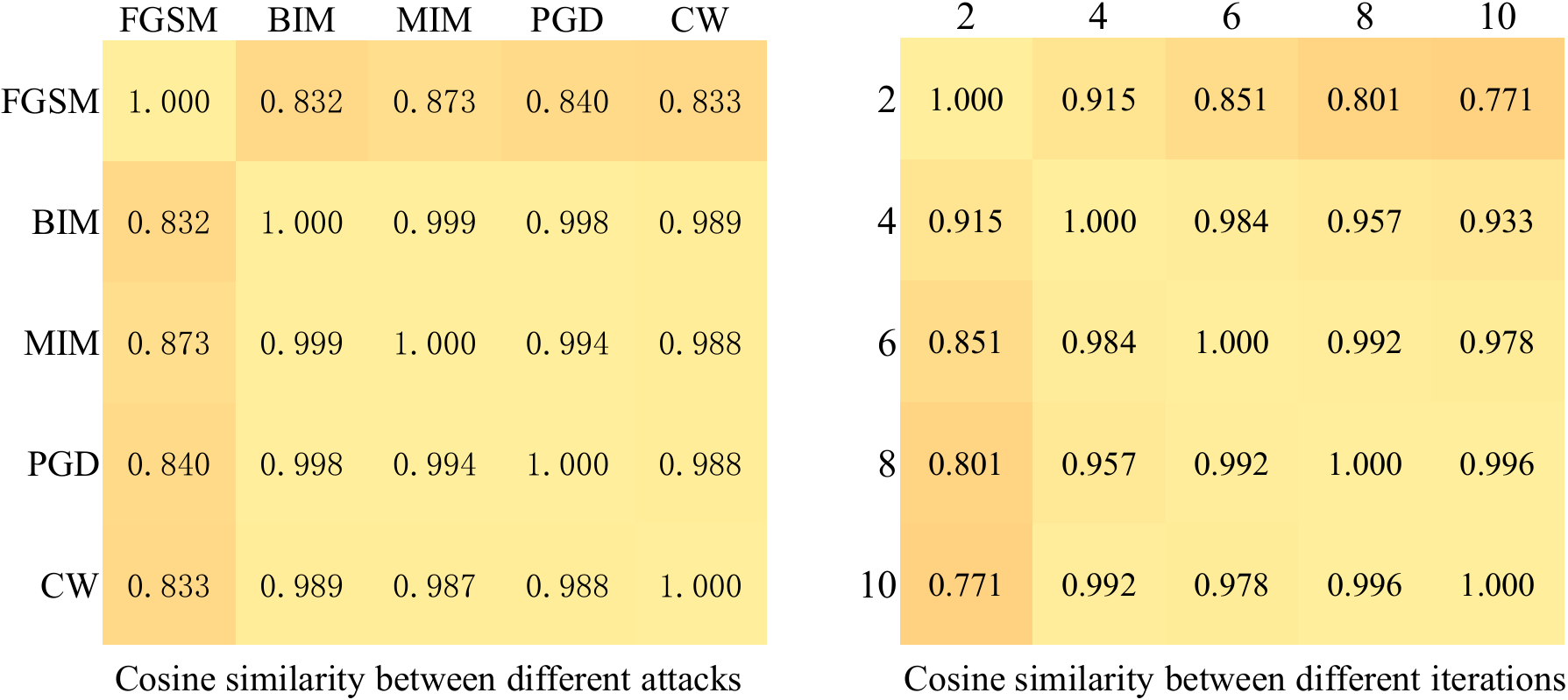}(b)
\end{center}
  \caption{(a) The similarity between the change values (under adversarial attack) from the penultimate layer and $(w_{i0} - w_{i1})$. (b)The similarity of the changes of features from the penultimate layer between different attacks and the different iterations from basic iterative method (BIM).}
\label{Fig:cos_weights}
\end{figure}

\textbf{Implication.} The partial derivative of cross-entropy loss $J$ with respect to the activation  value $z_i$ of $i$-th node in the penultimate layer is computed as:
\begin{equation}
    \nabla_{z_i} J(h(x), y_{1}) = (1 - p_1)(w_{i0} - w_{i1}),
\end{equation}
where $p_1$ denotes the probability of class 1 and $w_{ij}$ denotes the weight between $i$-th node in penultimate layer and $j$-th node in the last layer. Accordingly, the activation $z_i$ with larger $w_{i0} - w_{i1}$ will increase more (guided by the gradient) under attack. We plot the similarity between the value changes and $w_{i0} - w_{i1}$ in Fig.~\ref{Fig:cos_weights}(a). Similar conclusions can be derived 
for other attacks. As shown in Fig.~\ref{Fig:cos_weights}(b), \textbf{the features are pushed toward a similar direction during different iterations of different attacks}.

\section{Adversarial Attack with a Hierarchical Feature Constraint}
\label{Sec:Method}

We investigate the strategies that the attackers can leverage to bypass detection. Our key intuition is to exploit the vulnerability of medical features in the opposite way, i.e., pushing and hiding it in the distribution of normal medical features.


\bheading{Modeling the distribution of normal features.} We model the distribution of normal features using a Gaussian mixture model (GMM). 
\begin{equation}
    p(f^l(x)) = \sum_{k=1}^{K}\pi_k \mathcal{N}(f^l(x)|\mu_{k}, \Sigma_{k}),
\end{equation}
where $p$ is the probability density of sample $x$ in the target class $c$; $f^l(\cdot)$ denotes the feature of the $l^{th}$ activation layer; $\pi_k$ is the mixture coefficient subject to $\sum_{k=1}^{K}\pi_k=1$; $\mu_k$ and $\Sigma_k$ are the mean and covariance matrix of the $k$-th Gaussian component in the GMM. These parameters are trained by the expectation-maximization (EM) algorithm~\cite{EM} on the normal features belonging to the target class $c$.

\bheading{Hierarchical feature constraint (HFC).} We propose a hierarchical feature constraint as a simple-yet-effective add-on that can be applied to existing attacks to avoid being detected. First, we compute the log-likelihood of the AE feature under the distribution of normal features. Specifically, for a given input $x$, we separately compute the log-likelihood of an AE feature relative to each component and find the most likely Gaussian component:
\begin{equation}
k' = \mathop{\arg\max}\limits_k ~ \ln(\pi_k \mathcal{N}(f^l(x)|\mu_{k}, \Sigma_{k})). \label{eq:k'}
\end{equation}
To hide the adversarial representation, we maximize the log-likelihood of this chosen component. Conceptually, it encourages adversarial features in the high-density regions of the distribution of normal features. Then we encourage hiding in all DNN layers. The hierarchical feature constraint induces a loss ($J_{\mathrm{HFC}}$) that is formulated as Eq.~\ref{Eq:HFC}. It can be used as an add-on to existing attacks by directly adding $J_{\mathrm{HFC}}$ to any original attack object function $J_{original}$, i.e., $J = J_{original} + J_{\mathrm{HFC}}$.
\begin{equation}
    J_{\mathrm{HFC}} =  \sum^L_{l=1} \frac{\lambda^l}{2}(f^l(x) - \mu^l_{k'})^\top (\Sigma_{k'}^{l})^{-1} (f^l(x) - \mu^l_{k'}),
\label{Eq:HFC}
\end{equation}
where $\lambda^l$ weights the contribution of constraint in layer $l$. The pseudo-code of the adversarial attack boosted with HFC can be found in supplemental material.





\section{Experiments}


\subsection{Setup}
\label{Sec:setup}

\bheading{Datasets.} Following the literature~\cite{finlayson2018adversarial}, we use two public datasets on typical medical disease diagnosis tasks. The first one is the Kaggle Fundoscopy dataset~\cite{aptos} on the diabetic retinopathy (DR) classiﬁcation task, which consists of 3,663 fundus images labeled to one of the five levels from ``no DR'' to ``mid/moderate/severe/proliferate DR''. Following~\cite{finlayson2018adversarial}, we consider all fundoscopies with DR as the same class.  The other one is the Kaggle Chest X-Ray~\cite{CXR} dataset on the pneumonia classification task, which consists of 5,863 X-Ray images labeled with ``Pneumonia'' or ``Normal''. We split both datasets into three subsets: \textit{Train}, \textit{AdvTrain} and \textit{AdvTest}. For each dataset, we randomly select 80\% of the samples as \textit{Train set} to train the DNN classifier, and treat the left samples as the \textit{Test set}. The incorrectly classified (by the diagnosis network) test samples are discarded. Then we use 70\% of the samples (\textit{AdvTrain}) in the \textit{Test set} to train the adversarial detectors and evaluate their effectiveness with the remaining ones (\textit{AdvTest}).

\bheading{DNN models.} We choose the ResNet-50~\cite{resnet} and VGG-16~\cite{VGG} models pretrained on ImageNet. Both models achieve high Area Under Curve (AUC) scores on Fundoscopy and Chest X-Ray datasets: ResNet-50 achieves 99.5\% and 97.0\%, while VGG-16 achieves 99.3\% and 96.5\%, respectively.

\bheading{Adversarial attacks and detectors.} Following~\cite{ma2020understanding}, we choose three representative attacks, i.e., BIM, PGD and CW, against our models. For the adversarial detectors, we use kernel density (KD)~\cite{kde}, bayesian uncertainty (BU)~\cite{kde}, local intrinsic dimensionality (LID)~\cite{ma2018characterizing}, Mahalanobis distance (MAHA)~\cite{MAHA}, RBF-SVM~\cite{SaftyNet}, and deep neural network (DNN)~\cite{metzen2017detecting}. The parameters for KD, LID, BU and MAHA are set per the original papers. For KD, BU, and RBF-SVM, we extract features from the penultimate layers. For DNN, we train a classifier for each layer and ensemble by summing up their logits.

\bheading{Metrics.} We choose three metrics to evaluate the effectiveness of the adversarial detector and the proposed HFC bypassing method: 1) \textit{True positive rate at 90\% true negative rate (TPR@90):} The detector will drop 10\% of the normal samples to reject more adversarial attacks;  2) \textit{Area Under Curve (AUC) score}; 3) \textit{Adversarial accuracy (Adv. Acc):} The success rate of the targeted adversarial attack against diagnosis network.

\bheading{Hyperparameters.} We set\footnote{The hyperparameter analysis can be found in the supplementary material.} $K = 64$ and $1$ for Fundoscopy and Chest X-Ray datasets, respectively. For the $l^{th}$ layer, we compute the mean value of each channel separately, and set $\lambda_l$ to $\frac{1}{C_l}$, where $C_l$ is the number of channels. As tiny perturbations in medical images can cause drastic increase in the loss~\cite{ma2020understanding}, we set $\alpha = 0.02/256$ and $T = 2\epsilon / \alpha$.

\begin{table}[htbp]
\caption{The point-wise performances of HFC. The metrics scores on the \textit{left} and \textit{right} of the slash are the performances (\%) of the adversarial detectors under the attack \textit{without} and \textit{with} HFC, respectively. All of the attacks are evaluated on ResNet-50 with constraint $L_\infty=1/256$. }
\centering
\begin{tabular}{l|r|r|r|r|r|r}
\bottomrule \hline 
 \multirow{2}{*}{Fundoscopy} & \multicolumn{2}{c|}{BIM (Adv. Acc=99.5)} & \multicolumn{2}{c|}{PGD (Adv. Acc=99.5)} & \multicolumn{2}{c}{CW (Adv. Acc=99.5)} \\ \cline{2-7}  & \multicolumn{1}{c|}{AUC}         & \multicolumn{1}{c|}{TPR@90}        & \multicolumn{1}{c|}{AUC}         & \multicolumn{1}{c|}{TPR@90}       & \multicolumn{1}{c|}{AUC}         & \multicolumn{1}{c}{TPR@90}       \\ 
\hline
KD         & 99.0 / 74.2  & 96.8 / 20.5  & 99.4 / 73.4  & 98.6 / 13.2  & 99.5 / 74.7  & 99.1 / 19.6 \\
MAHA          & 99.6 / ~6.4   & 99.5 / ~0.0     & 100 / ~4.2    & 100 / ~0.0      & 99.8 / 33.0  & 99.5 / ~0.0    \\
LID        & 99.8 / 78.3  & 100 / 40.6   & 99.6 / 73.2  & 98.6 / 35.5  & 98.8 / 73.4  & 97.7 / 33.3 \\
SVM        & 99.5 / 28.6  & 99.1 / ~0.0     & 99.8 / 23.1  & 99.5 / ~0.0     & 99.8 / 27.0  & 99.5 / ~0.0    \\
DNN         & 100 / 60.0     & 100 / 12.8   & 100 / 58.6   & 100 / ~8.2    & 100 / 62.6   & 100 / 15.1  \\
BU         & 58.9 / 37.4  & 9.1 / ~0.0      & 61.9 / 35.9  & 9.1 / ~0.0      & 93.0 / 32.8  & 73.1 / ~5.0   \\
\hline
 \multirow{2}{*}{Chest X-Ray} & \multicolumn{2}{c|}{BIM (Adv. Acc=90.9)} & \multicolumn{2}{c|}{PGD (Adv. Acc=90.9)} & \multicolumn{2}{c}{CW (Adv. Acc=98.9)} \\ \cline{2-7} & \multicolumn{1}{c|}{AUC}         & \multicolumn{1}{c|}{TPR@90}        & \multicolumn{1}{c|}{AUC}         & \multicolumn{1}{c|}{TPR@90}        & \multicolumn{1}{c|}{AUC}         & \multicolumn{1}{c}{TPR@90}     \\ 
\hline
KD            & 100 / 73.1    & 100 / ~6.8  & 100 / 82.3    & 100 / 50.5  & 99.2 / 71.5   & 98.4 / 15.7    \\
MAHA              & 100 / ~0.0   & 100 / ~0.0    & 100 / ~0.0  & 100 / ~0.0     & 100 / 22.4          & 100 / ~0.0     \\
LID             & 100 / 48.6    & 100 / ~1.8   & 100 / 49.1    & 100 / ~1.5   & 99.2 / 64.5     & 98.4 / 14.4    \\
SVM       & 100 / 16.7    & 100 / ~6.9     & 100 / ~5.8    & 100 / ~0.0     & 100 / 21.2  & 100 / ~0.0 \\
DNN                & 100 / 31.8    & 100 / ~0.7   & 100 / 33.7    & 100 / ~0.0     & 100 / 61.6    & 100 / ~5.2     \\
BU                & 49.9 / 26.1   & 19.2 / ~0.0    & 49.2 / 26.2   & 22.7 / ~0.0    & 98.3 / 26.2      & 94.8 / ~0.0     \\
\hline \toprule 
\end{tabular}
\label{Table:pointwise}
\end{table}

\begin{figure}[h]
\begin{center}
   \includegraphics[width=0.95\linewidth]{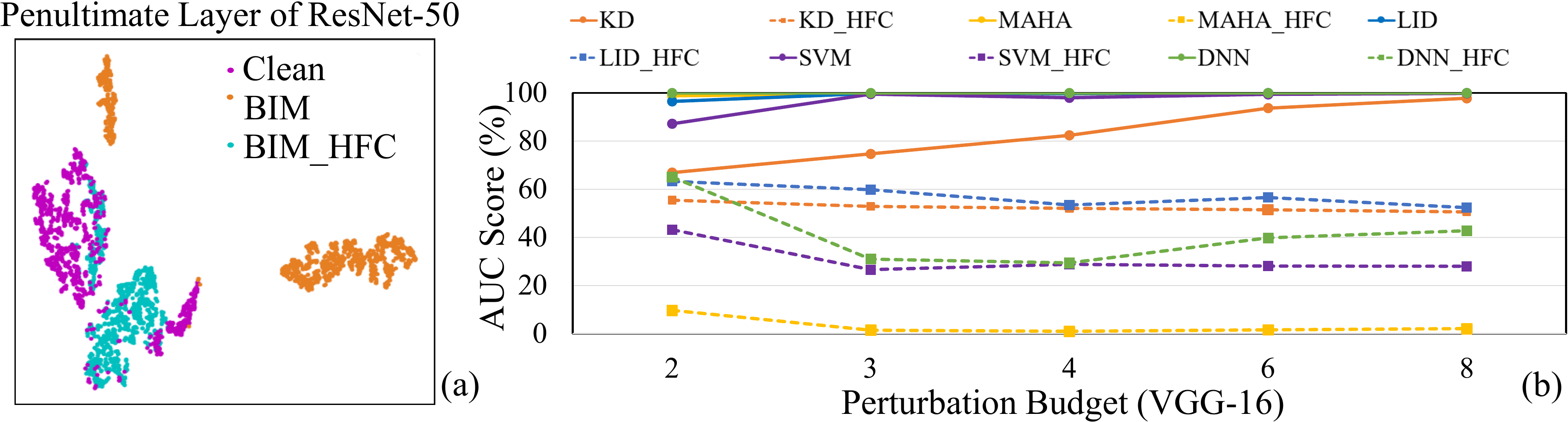}
\end{center}
   \caption{(a)Visualization of 2D t-SNE of clean and adversarial features generated from BIM and HFC, extracted from ResNet-50 on Chest X-Ray. (b) The AUC scores in the solid lines and dotted lines are the performances of the adversarial detectors against BIM \textit{with} and \textit{without} HFC.}
\label{Fig:Pertubation}
\end{figure}

\subsection{Experimental Results}
\label{Sec:experiments_main}

\bheading{Bypassing adversarial detectors.} We compare the performances of the existing adversarial attacks \textit{with} and \textit{without} HFC on different DNN classifiers, datasets, and perturbation constraints. The visualization in Fig.~\ref{Fig:Pertubation}(a) shows that HFC successfully moves the AE feature (orange) from outlier regions to the regions (cyan) where normal features (purple) reside. Quantitative results in Table~\ref{Table:pointwise} show that the HFC boosts all evaluated adversarial attacks to bypass all evaluated detectors simultaneously, with high accuracy and a small perturbation constraint $L_\infty=1/256$. Furthermore, as shown in Fig.~\ref{Fig:Pertubation}(b), a larger perturbation budget provides HFC more room to manipulate the representations, compromising the detectors more drastically (as the dotted lines decrease).

\bheading{Comparison to other adaptive attacks.} We compare the proposed HFC to other adaptive attacks designed to manipulate the deep representations and bypass the detectors: 1) Generate AEs with internal representation similar to a random guide image~\cite{feature_iclr}; 2) Choose a guide image with its representation closest to the input~\cite{feature_iclr}; 3) Minimize the loss terms of KDE and cross-entropy simultaneously~\cite{CW_ten}; 4) Minimize the loss term of LID and cross-entropy at the same time~\cite{CW_bpda}. As shown in Table~\ref{Table:sota}, our proposed HFC bypasses the five detectors and greatly outperforms other competitive adaptive attacks. 

\begin{table}[htp]
\caption{AUC scores (\%) of the proposed HFC and other adaptive attacks. The AEs are generated on ResNet-50 under constraint $L_\infty=1/256$.}
\centering
\begin{tabular}{l|rrrrr|l|rrrrr}
\bottomrule \hline
Fundoscopy & KD   & MAHA & LID  & SVM  & DNN   & Chest X-Ray & KD   & MAHA & LID  & SVM  & DNN  \\
\hline
Random     & 75.1 & 86.1 & 91.7 & 48.2 & 93.7  &  Random         & 77.0   & 64.0   & 91.0 & 13.0   & 79.3  \\
Closest    & 77.0   & 64.0   & 91.0 & 13.0   & 79.3 & Closest & 80.1 & 38.3 & 71.3 & 9.9  & 87.7 \\
KDE        & 51.6 & 86.5 & 90.9 & 45.3 & 95.0   & KDE                      & 58.2 & 66.9 & 71.7 & 15.3 & 95.6\\
LID        & 87.6 & 85.4 & 93.4 & 61.2 & 96.2  &  LID                      & 84.0 & 66.6 & 77.1 & 28.6 & 96.6 \\
HFC   & 74.2 & 6.4  & 78.3 & 28.6 & 60.0  & HFC                  & 70.8 & 0.0 & 53.6 & 16.7 & 32.6 \\
\hline \toprule
\end{tabular}
\label{Table:sota}
\end{table}




\bheading{Semi-white-box attack.} We evaluate the proposed method in a more difficult scenario: only the architecture information of DNN models is available. The attacker tries to confuse the victim model and bypass its adversarial detectors simultaneously, without knowing model parameters. The results in Table~\ref{Table:gray-box} show that our HFC can help BIM compromise most of the detectors and manipulate the DNN models,  which poses more disturbing concerns to the safety of DNN-based diagnosis networks.

\begin{table}[H]
\caption{The performance of BIM \textit{w/} and \textit{w/o} HFC under the semi-white-box setting. .  The  AEs  are generated under constraint $L_\infty=4/256$. AUC scores (\%) are used as metrics.}
\centering
\begin{tabular}{l|l|rrrrrrrr}
\bottomrule \hline
DNN Model & Fundoscopy                 & KD                   & MAHA                 & LID                  & SVM                  & DNN                  & Adv. Acc              \\
\hline
 \multirow{2}{*}{ResNet-50} & BIM   & 98.7                 & 100.0                 & 99.5                 & 92.1                 & 100.0                  & 83.2                 \\
 & BIM\_HFC  & 78.0                   & 9.1                  & 68.0                   & 16.9                 & 43.8                 & 68.2                 \\
  \cline{1-2} \multirow{2}{*}{VGG-16} & BIM     & 72.3                 & 89.6                 & 81.5                 & 48.1                 & 95.2                 & 88.6                 \\
& BIM\_HFC      & 50.9                 & 18.2                 & 64.6                 & 28.8                 & 16.7                 & 73.2                 \\
\hline \toprule
\end{tabular}
\vspace{-1mm}
\label{Table:gray-box}
\end{table}

\section{Conclusion}
In this paper, we investigate the characteristics of medical AEs in feature space. A stress test is performed to reveal the greater vulnerability of medical image features, compared to the natural images. Then we theoretically prove that existing adversarial attacks tend to alter the vulnerable features in a fixed direction. As a result, the adversarial features become outliers and easy to detect. However, an attacker can exploit this vulnerability to hide the adversarial features. We propose a novel hierarchical feature constraint (HFC), a simple-yet-effective add-on that can be applied to existing attacks, to avoid AEs from being detected. Extensive experiments validate the effectiveness of HFC, which also significantly outperforms other adaptive attacks. It reveals the limitation of the current methods for detecting medical AEs in the feature space. We hope it can inspire more defenses in future work.

\newpage



\end{document}


\title{A Hierarchical Feature Constraint to Camouflage Medical Adversarial Attacks \\
\textit{Supplementary Material}}

\maketitle

\textbf{Theorem 1.} \textit{Consider a binary disease diagnosis network and its representations from the penultimate layer, the directions of the corresponding gradients are fixed during during each iteration under adversarial attack.\footnote{As a representative, we use adversarial attack to convert the prediction of diagnosis network from 0 to 1.}}

\textbf{Proof.} $h$ denotes the disease diagnosis network; $z_i$ denotes the activation output of $i$-th node (neuron) in the penultimate layer of a neural network; $w_{ij}$ denotes the weight from $i$-th node to $j$-th node of the last layer; $l_i$ is the output of neural network in $i$-th channel; $p_i$ is the probability of the input being $i$-th class. For simplicity, we ignore the bias parameters and activation functions in the dense layer, which does not affect the conclusion. Formally, the cross-entropy loss can be defined as:
\begin{equation}
    J_{CE} = -\sum_k y_klog(p_k),
\end{equation}
and the Softmax function is:
\begin{equation}
    p_i = \frac{e^{l_i}}{\sum_k e^{l_k}}.
\end{equation}

According to the chain rule, we compute the partial derivative of $J$ with respect to $l_i$:
\begin{equation}
\label{eq:gradient_J_l}
    \nabla_{l_i} J(h(x)) = p_i - y_i.
\end{equation}
Hence,
\begin{equation}
    \begin{split}
        \nabla_{z_i} J(h(x)) &= \nabla_{l_0} J \times \nabla_{z_0} l_0 + \nabla_{l_1} J \times \nabla_{z_0} l_1 \\
    &= (p_0 - y_0)w_{i0} + (p_1 - y_1)w_{i1},
    \end{split}
\end{equation}
the goal of adversarial attack is to invert the prediction from 0 to 1, i.e., $y_0=0, y_1=1$, so we can derive the partial derivative on $h_n^{i}$:
\begin{equation}
    \begin{split}
        \nabla_{z_i} J(h(x), y_{1}) &= p_0w_{i0} + (p_1 - 1)w_{i1} \\
        &= (1 - p_1)(w_{i0} - w_{i1}),
    \end{split}
\end{equation}
where $1 - p_1>0$ and $w_n^{i0} - w_n^{i1}$ is constant, so the partial derivative on $z_n^{i}$ will keep in the same direction, as is claimed in Sec. 3.

\begin{table}[htp]
\caption{The AUC scores (\%) under different number of components $K$ of GMM on Fundoscopy (Fun.) and Chest X-Ray (CXR).}
\vspace{0.1cm}
\centering
\small
\scalebox{0.8}{
\begin{tabular}{l|rrrrrrrr|l|rrrrrrrr}
\bottomrule \hline
  $K (Fun.)$  & 1    & 2    & 4    & 8    & 16   & 32   & 64   & 128 & $K (CXR)$  & 1    & 2    & 4    & 8    & 16   & 32   & 64   & 128  \\
\hline
KD  & 75.8 & 78.5 & 77.9 & 77.8 & 73.6 & 73.3 & 74.2 & 74.4 & KD  & 70.3 & 71.0 & 66.5 & 64.9 & 65.6 & 63.7 & 61.7 & 62.8\\
MAHA & 5.9 & 5.6 & 6.2 & 7.1 & 8.2 & 8.6 & 7.6 & 5.9 & MAHA & 0.0 & 0.0 & 0.0 & 0.0 & 0.0 & 0.0 & 0.0 & 0.0\\
LID & 81.7 & 82.4 & 83.0 & 84.2 & 83.2 & 80.7 & 78.3 & 78.7 & LID & 48.3 & 52.3 & 55.5& 58.8 & 49.8 & 49.0 & 48.5 & 49.4\\
SVM & 52.3 & 45.7 & 40.4 & 34.0 & 31.4 & 34.5 & 35.5 & 40.3 & SVM & 20.6 & 18.7 & 17.8 & 14.4 & 12.3 & 14.3 & 13.6 & 13.5\\
DNN & 61.4 & 60.7 & 62.6 & 64.3 & 64.4 & 63.5 & 63.4 & 64.9 & DNN & 31.8 & 32.7 & 38.6 & 41.9 & 40.2 & 39.5 & 38.9 & 37.9\\
BU & 55.5 & 52.2 & 49.7 & 45.2 & 43.3 & 42.6 & 43.8 & 42.6 & BU & 44.2 & 35.6 & 27.8 & 22.2 & 25.8 & 26.2 & 25.6 & 26.3\\
\hline \toprule
\end{tabular}
}
\vspace{-5mm}
\label{Table:k2}
\end{table}

\renewcommand{\algorithmicrequire}{ \textbf{Input:}} 
\renewcommand{\algorithmicensure}{ \textbf{Output:}} 

\begin{algorithm}[h]
\caption{Adversarial attack with hierarchical feature constraint}
\label{algo1}
\begin{algorithmic}[1]
\REQUIRE~~\\
Training set $(x^i, y^i)\in \mathcal{D}$, input image $x$, target class $c$, Model $h$, iteration $T$, step size $\alpha$

\ENSURE ~~\\
Adversarial image $x^*$ with $||x^* - x||_\infty \leq \epsilon$

\FOR{each DNN's layer $l=1$ to $L$}
\STATE Initialize mean $\mu_{k}^l$ and covariance $\Sigma_{k}^l$ of each component, where $k \in \{1, ..., K\}$
\STATE Train the GMM using the EM algorithm with $f^l(x^i)$,  where $x^i \in \mathcal{D}$ with $y^i = c$
\ENDFOR
\STATE $x_0^* = x$
\FOR{$t=0$ to $T$}
\STATE $J_{original} =  J_{\mathrm{HFC}} + J_{original}$,
\STATE $x_{t+1}^* = \mathrm{clip}_\epsilon( x_t^* - \alpha \cdot \mathrm{sign}(\nabla_x (J)))$
\ENDFOR
\RETURN $x^*$
\end{algorithmic}
\end{algorithm}

\begin{figure*}[htp]
    \centering
    \includegraphics[width=1\linewidth]{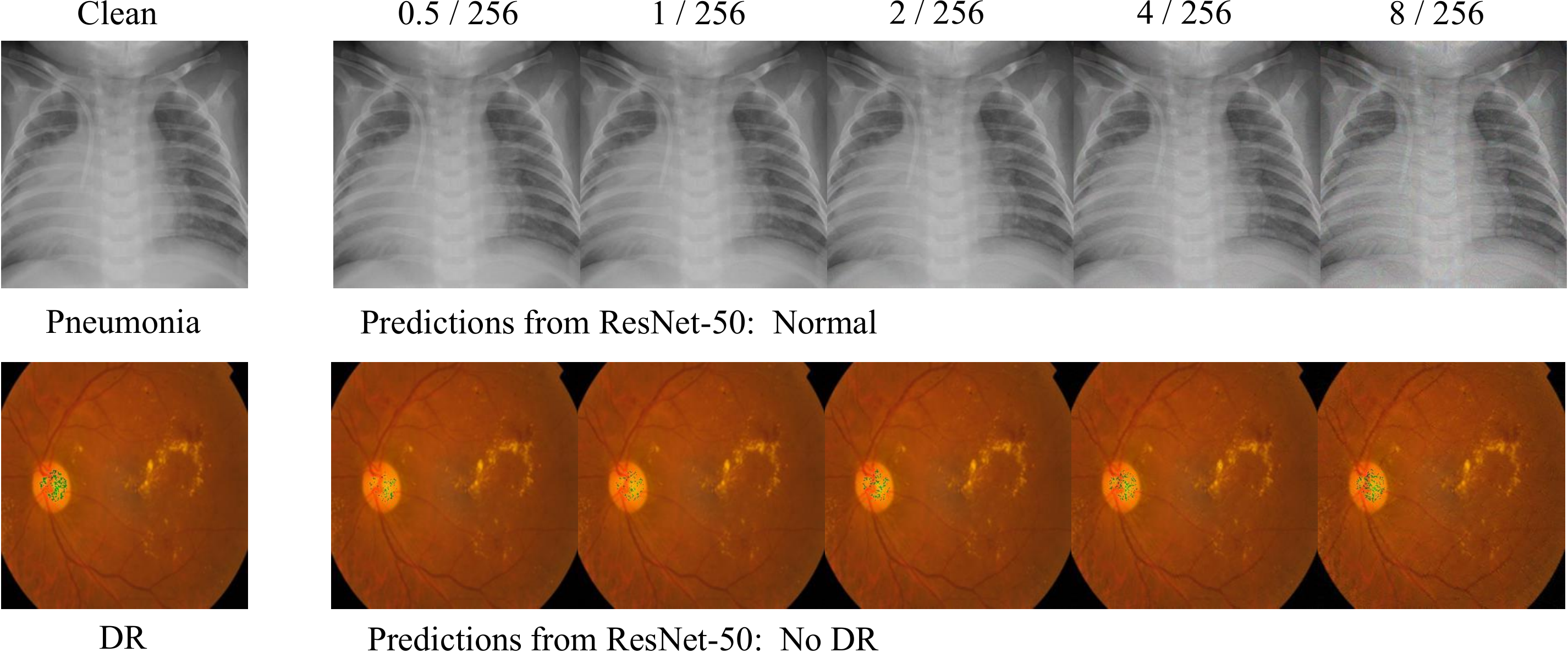}
    \caption{Visualization for the clean and adversarial examples. The numbers above the pictures represent the $L_\infty$-norm of the perturbations.}
    \label{fig:visualize}
\end{figure*}